\documentclass{article}
\usepackage{dirtree} 
\usepackage{array}
\usepackage{indentfirst} 
\usepackage{float} 
\usepackage{url}
\usepackage{cite}
\usepackage[english]{babel}

\usepackage[letterpaper,top=2cm,bottom=2cm,left=3cm,right=3cm,marginparwidth=1.75cm]{geometry}

\usepackage{amsmath}
\usepackage{graphicx}
\usepackage[colorlinks=true, allcolors=blue]{hyperref}

\title{A Multi-Person Video Dataset Annotation Method of Spatio-Temporally Actions}
\author{Fan Yang \\Chengdu Neusoft Institute Of Information\vspace{-2cm}
\thanks{ yang.fan@nsu.edu.cn} 
}

\begin{document}
\date{}
\maketitle

\begin{abstract}
Spatio-temporal action detection is an important and challenging problem in video understanding. However, the application of the existing large-scale spatio-temporal action datasets in specific fields is limited, and there is currently no public tool for making spatio-temporal action datasets, it takes a lot of time and effort for researchers to customize the spatio-temporal action datasets, so we propose a multi-Person video dataset Annotation Method of spatio-temporally actions.First, we use ffmpeg to crop the videos and frame the videos; then use yolov5 to detect human in the video frame, and then use deep sort to detect the ID of the human in the video frame. By processing the detection results of yolov5 and deep sort, we can get the annotation file of the spatio-temporal action dataset to complete the work of customizing the spatio-temporal action dataset.Code will be made publicly available at \url{https://github.com/Whiffe/Custom-ava-dataset_Custom-Spatio-Temporally-Action-Video-Dataset}
\end{abstract}

\section{Introduction}

\begin{figure}[H]
\centering
\includegraphics[width=0.9\textwidth]{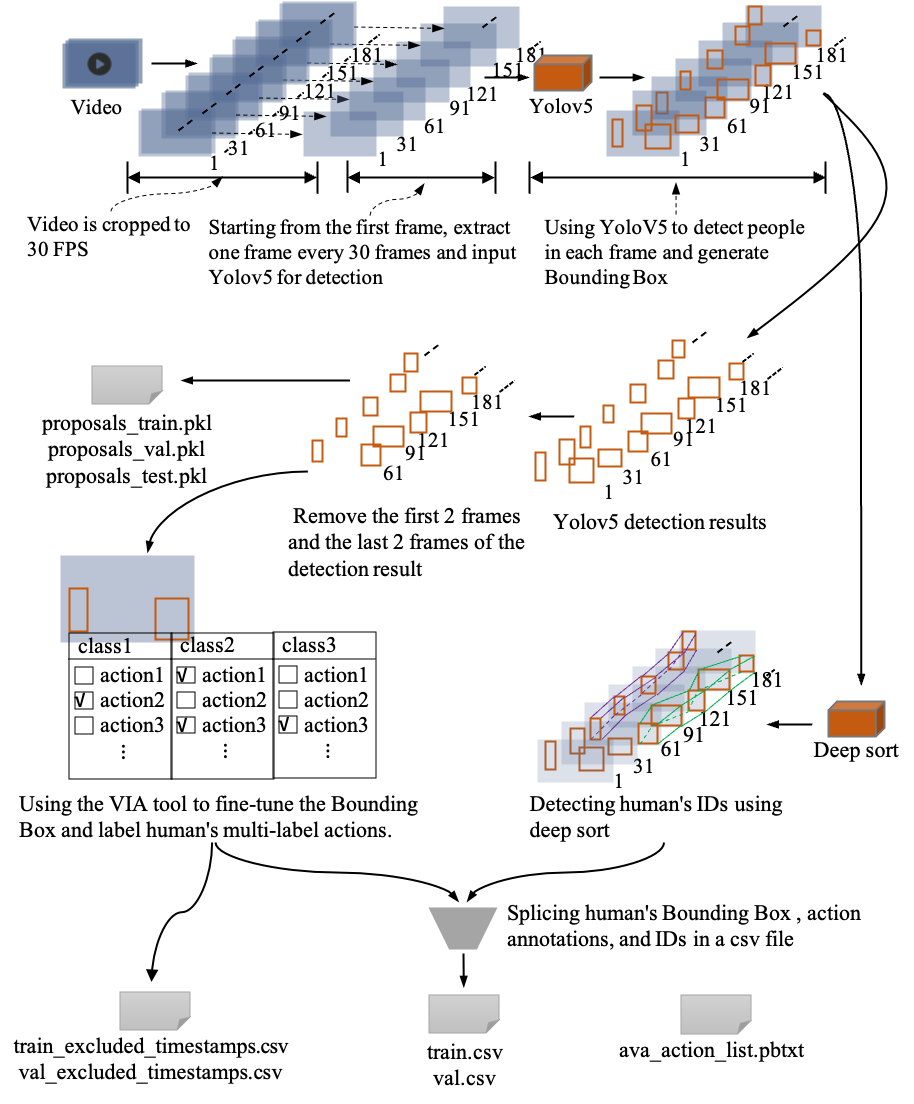}
\caption{Production process of Spatio-Temporally Actions datasets}\label{ActionsDatasetsProcess}
\end{figure}

The rapid development of video understanding\cite{zhu2020comprehensive}, has led to significant development in the field of spatio-temporal human action detection, such as ACRN\cite{sun2018actor}, SlowOnly, SlowFast\cite{Feichtenhofer_2019_ICCV}, LFB\cite{Wu_2019_CVPR}, etc. Currently, remarkable progress has been made in building increasingly complex and realistic spatio-temporal datasets, such as the UCF101-24\cite{Soomro2012UCF101AD} , JHMDB\cite{Jhuang:ICCV:2013}, AVA\cite{gu2018ava} and MultiSports\cite{li2021multisports} datasets. However, the application in specific scenarios requires custom spatio-temporal datasets. The production workload of  spatio-temporal datasets is large, and there is no public production tool for spatio-temporal datasets, therefore, this paper proposes a multi-Person video dataset Annotation Method of
spatio-temporally actions.Method like figure \ref{ActionsDatasetsProcess}

\section{Method of making Spatio-Temporally Actions datasets}

\subsection{The details of Spatio-Temporally actions datasets annotation format}
\begin{figure}[H]
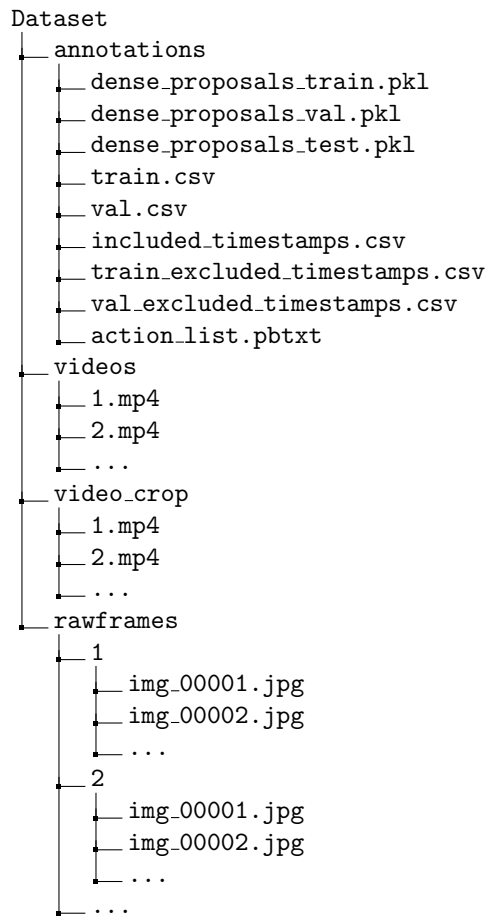

\dirtree{%
    .1 Dataset.
    .2 annotations.
    .3 dense\_proposals\_train.pkl.
    .3 dense\_proposals\_val.pkl.
    .3 dense\_proposals\_test.pkl.
    .3 train.csv.
    .3 val.csv.
    .3 included\_timestamps.csv.
    .3 train\_excluded\_timestamps.csv.
    .3 val\_excluded\_timestamps.csv.
    .3 action\_list.pbtxt.
    .2 videos.
    .3 1.mp4.
    .3 2.mp4.
    .3 ....   
    .2 video\_crop.
    .3 1.mp4.
    .3 2.mp4.
    .3 ....
    .2  rawframes.
    .3 1.
    .4 img\_00001.jpg.
    .4 img\_00002.jpg.
    .4 ....
    .3 2.
    .4 img\_00001.jpg.
    .4 img\_00002.jpg.
    .4 ....
    .3 ....
}
\caption{\label{fig:frog}Dataset‘s folder structure}
\end{figure}

The spatio-temporally actions dataset consists of 4 parts: original videos, cropped videos, video frames, and annotation files.

Original videos: We collect relevant original videos based on custom action categories. Such as student classroom, subway station, factory and other scene videos. 

Cropped videos: We crop the original videos into videos of the same length, such as 15 seconds or 15 minutes. 

Rawframes: We use the video cropping tool (ffmpeg) to crop the video to 30 frames per second.

Annotations: This part is the core of this paper. The details of each annotation file are as follows\cite{2020mmaction2} :

1) The proposals generated by human detectors 

\begin{table}[H]
	\centering
	\caption{The example of dense\_proposals\_[train/val/test].pkl}
	\label{tab:1}  
	\begin{tabular}{cc}
		\hline\noalign{\smallskip}	
		\{ “1,0002”: [  \\
		 & [0.670  0.264 0.928 0.835 0.330391] \\
		 & [0.265  0.164 0.942 0.866 0.990325] \\
		 & [0.000  0.259 0.180  0.886 0.229031] \\
		 & [0.126 0.352 0.403 0.848 0.985255] \\
		 & [0.142 0.214 0.634 0.886 0.053679]] \\
		 “1,0003”: [  \\
		 & [0.647 0.254 0.951 0.850  0.303808] \\
		 & [0.270  0.173 0.946 0.869 0.991948] \\
		 & [0.000  0.263 0.187 0.897 0.324760] \\
		 & ... \\
		... \\
		\noalign{\smallskip}\hline
	\end{tabular}
\end{table}

Dense\_proposals\_[train/val/test].pkl are human proposals generated by a human detector. They are used in training, validation and testing respectively. Take table \ref{tab:1}  dens\_proposals\_train.pkl as an example. The key consists of the videoID and the timestamp. For example, the key “1,0002“ means the values are the detection results for the frame at the $2_{nd}$ second in the video 1. The values in the dictionary are numpy arrays with shape $N \times 5$, N is the number of detected human bounding boxes in the corresponding frame. The format of bounding box is $[ x_1 , y_1 , x_ 2 , y_2 , score] $, $0 \leq x_1 , y_1 , x_2 , w_2, score \leq 1$, $(x_1,y_1)$ indicates the top-left corner of the bounding box, $(x_2,y_2)$ indicates the bottom-right corner of the bounding box; $(0,0)$ indicates the top-left corner of the image, while $(1,1)$ indicates the bottom-right corner of the image.

2) The ground-truth labels for spatio-temporal action detection
\begin{table}[H]
	\centering
	\caption{The example of [train/val].csv}
	\label{tab:2}  
	\begin{tabular}{cc}
		\hline\noalign{\smallskip}	
		[  \\
		 & ["1","2","0.114","0.353","0.401”,"0.848","1","0"] \\
		 & ["1","2","0.114","0.353","0.401”,"0.848","6","0"] \\
		 & ["1","2","0.253","0.177","0.932","0.860","1","1"] \\
		 & ["1","3","0.119","0.355","0.405","0.845","1","0"] \\
		 & ["1","3","0.232","0.201","0.908","0.861","1","1"] \\
		 & ["1","3","0.232","0.201","0.908","0.861","4","1"] \\
		 & … \\
		 & ["1","13","0.172","0.133","0.691","0.748","1","13"] \\
		 & ["1","13","0.172","0.133","0.691","0.748","2","13"] \\
		 & ["2","1","0.189","0.476","0.474","0.859","1","0"] \\
		 & ["2","1","0.189","0.476","0.474","0.859","4","0"] \\
		 & ["2","1","0.189","0.476","0.474","0.859","5","0"] \\
		 & … \\
		\noalign{\smallskip}\hline
	\end{tabular}
\end{table}

In the annotation folder, [train/val].csv are ground-truth labels for spatio-temporal action detection, which are used during training \& validation.  Take table \ref{tab:2}  train.csv as an example, it is a csv file with annotation lines, each line is a human instance in one frame. For example, the first line in train.csv is ["1","2","0.114","0.353","0.401”,"0.848","1","0"]:the first two items "1" and "2" indicate that it corresponds to the $2\_nd$ second in the video 1.The next four items $([0.114,0.353,0.401,0.848])$ indicates the location of the bounding box, the bbox format is the same as human proposals. The next item 1 is the action label. The last item 0 is the ID of this bounding box.

3) Included timestamps

\begin{table}[H]
	\centering
	\caption{The example of included\_timestamps.csv}
	\label{tab:IncludedTimestamps}  
	\begin{tabular}{cc}
		\hline\noalign{\smallskip}	
		["002" "003" "004" "005"..."" "n-1" "n"] \\
		\noalign{\smallskip}\hline
	\end{tabular}
\end{table}
As shown in Table \ref{tab:IncludedTimestamps}, included\_timestamps.csv represents the sequence of video frames used in training  and validation.

4) Excluded timestamps

\begin{table}[H]
	\centering
	\caption{The example of [train/val]\_excludes\_timestamps.csv}
	\label{tab:3}  
	\begin{tabular}{cc}
		\hline\noalign{\smallskip}	
		["13","072"	"13","115"	"13","116"	"30","090"	"30","091"	"57","081"    ...] \\
		\noalign{\smallskip}\hline
	\end{tabular}
\end{table}

[train/val]\_excludes\_timestamps.csv contains excluded timestamps which are not used during training or validation. The format is video\_id, second\_idx. Such as ["13", "072"] indicates that the $72_{nd}$ frame in video 13 are not used in training or validation.

5) Label map

\begin{table}[H]
	\centering
	\caption{The example of action\_list.pbtxt}
	\label{tab:4}  
	\begin{tabular}{cc}
		\hline\noalign{\smallskip}	
        [  \\
        & item\{name: "action1"   id: 1\} \\
        & item\{name: "action2"   id: 2\} \\
        & item\{name: "action3"   id: 3\} \\
        & ... \\
        & item\{name: "actionn"   id: n\} \\
        ]  \\
		\noalign{\smallskip}\hline
	\end{tabular}
\end{table}

Action\_list.pbtxt contains the label map of the dataset, which maps the action name to the label index.

\subsection{Video selection, cropping and frame extraction}

First, we need to select a sufficient number of videos, and the video content is a specific scene, such as student classroom, subway station, factory and other scene videos.Next we crop the original videos into videos of the same length, such as 15 seconds or 15 minutes. Then we use the video cropping tool (ffmpeg) to crop the video to 30 frames per second.For example, a 15-second video will generate 451 video frames $(15 \times 30 + 1)$.

\subsection{YoloV5 detection}\label{YoloV5Detection}

We use the Yolov5 detection algorithm to detect the of bounding box of the human in the video frames, of course, we can also use other detection algorithms, such as Faster Rcnn, Yolov3, etc. It should be noted that not all video frames we need to detect, we only detect the $(30 \times n + 1)th$ video frame, such as 1,31,61,91....

\subsection{Generation of proposals\_[train/val/test].pkl file and human ID}\label{proposalsAndHumanID}

In \ref{YoloV5Detection} we have obtained bounding box for human in video frames. The detection results will be used in two ways, the first one is used to make dense proposals\_[train/val/test].pkl file, the second feeds the result and its corresponding video frame into Deep sort\cite{wojke2017simple} to generate the person's ID. When making proposals\_[train/val/test].pkl, we remove the first 2 frames and the last 2 frames in the detection results. When detecting a human's ID, since deep sort needs at least 3 video frames to judge, we do not remove it compared to proposals\_[train/val/test].pkl.

\subsection{Action annotation with VIA}\label{VIAAnnotation}

\begin{figure}[H]
\centering
\includegraphics[width=0.9\textwidth]{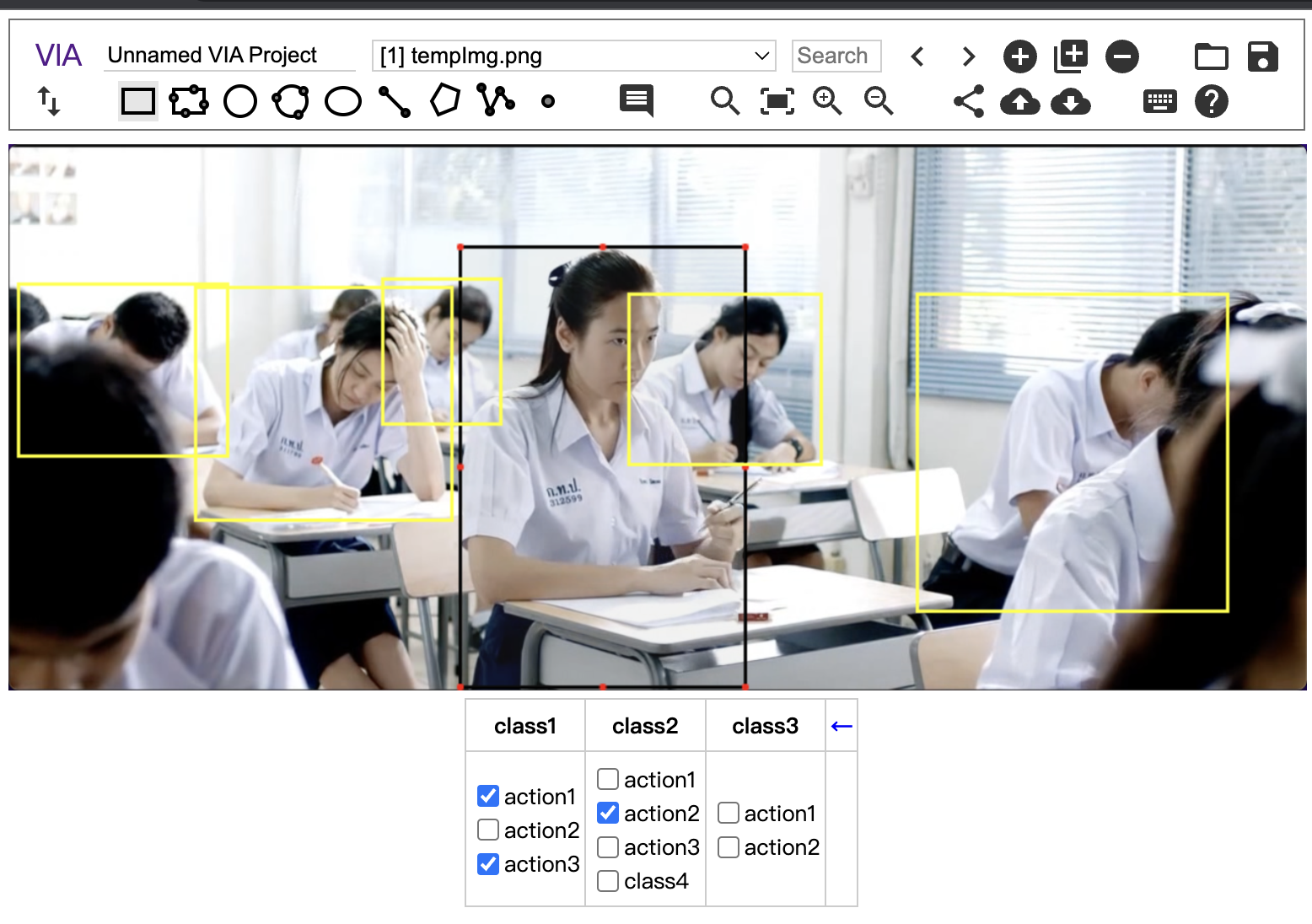}
\caption{VIA interface for action annotation}\label{viaAnnatationImg}

\end{figure}

We choose VIA tool for the labeling of multi-label behavior. The interface and labeling details of VIA are shown in figure \ref{viaAnnatationImg}, the rectangles in the figure \ref{viaAnnatationImg} is the  bounding box of the human read from proposals\_[train/val/test].pkl, we need to fine-tune the rectangle so that the rectangle can just cover the person, We also need to remove overlapping rectangles and add rectangles that are not detected.Finally, we annotate each rectangle with multi-label 
action.

\subsection{Generation of train/val.csv file}
In \ref{proposalsAndHumanID}, \ref{VIAAnnotation}, we have got: videoID, timestamp, bounding box, action, human IDs, so we combine the results of proposals\_[train/val/test].pkl and human IDs(from deep sort) to get [train/val].csv

\subsection{Generation of other annotation files}

Before we frame the video, we determine included\_timestamps.csv

Before using the VIA tool to label, we have determined the action classes and generated action\_list.pbtxt. 

When using via for annotation, we select frames which are not used during training or validation and write them into [train/val]\_excludes\_timestamps.csv

\bibliographystyle{IEEEtran}
\bibliography{citepaper}

\end{document}